\documentclass{article}
\usepackage{amsthm}

\usepackage[preprint,nonatbib]{neurips_2026}            

\usepackage{tabularx}
\usepackage{array}
\usepackage{makecell}

\newcolumntype{L}{>{\raggedright\arraybackslash}X}
\newcolumntype{C}[1]{>{\centering\arraybackslash}p{#1}}

\usepackage[T1]{fontenc}
\usepackage{microtype}

\newtheorem{theorem}{Theorem}[section]

\newtheorem{proposition}[theorem]{Proposition}

\usepackage{booktabs} 
\usepackage{multirow} 
\usepackage{adjustbox} 
\usepackage{array} 
\usepackage{arydshln} 

\usepackage[backend=biber, maxnames=1]{biblatex}
\usepackage{array}
\usepackage{amsmath}
\usepackage{amsfonts}
\usepackage{amssymb}
\usepackage{amsxtra}
\usepackage{amsthm}
\usepackage{mathrsfs}
\usepackage{mathtools}
\usepackage{bm}
\usepackage{cancel}
\usepackage{yhmath}
\usepackage{nicefrac}

\usepackage[table]{xcolor}
\usepackage{graphicx}
\usepackage{enumitem}
\usepackage{booktabs}
\usepackage{subcaption}
\usepackage{float}
\usepackage{tikz}
\usepackage{siunitx}
\usepackage{algorithm}
\usepackage{algpseudocode}

\usepackage[colorlinks=true, bookmarksopen=true]{hyperref}
\usepackage{url}

\definecolor{darkblue}{rgb}{0.0,0.2,0.6}

\theoremstyle{plain}

\theoremstyle{definition}

\theoremstyle{remark}

\title{Random-Effects Algorithm for Random Objects in Metric Spaces}

\author{%
  Marcos Matabuena\thanks{Corresponding author.} \\
   Mohamed bin Zayed University of Artificial Intelligence \\
  \texttt{Marcos.Matabuena@mbzuai.ac.ae} \\
  \And
  Mateo Cámara\thanks{MM and MC share first authorship.} \\
  Signal Processing Applications Group,
Universidad Politécnica de Madrid  \\
  \texttt{mateo.camara@upm.es} \\
}

\begin{document}

\maketitle
\begin{abstract}
Across many scientific disciplines, multiple observations are collected from the same experimental units, and in modern datasets these observations often arise as non-Euclidean random objects. In such settings, the incorporation of random effects is a critical modeling step for efficient estimation and personalized prediction. Although mixed-effects models are well established for scalar outcomes and, more recently, for functional data in Hilbert spaces, general random-effects frameworks for objects in metric spaces remain underdeveloped. In this paper, we propose a nonlinear Fréchet-based algorithm for random-effects modeling of arbitrary random objects defined on a metric space. Using M-estimation theory, we establish conditions under which the proposed metric-space prediction target is consistently estimated under a working random-effects formulation. We then evaluate the empirical performance of the proposed method using both synthetic data and digital health datasets that require practical tools for analyzing random objects in metric spaces, such as multivariate probability distributions and random graphs. We show that, although our method is developed beyond Hilbert spaces, it can outperform existing Hilbert space-based methods.
\end{abstract}


\noindent%
{\it Keywords:} random effects; metric spaces; digital health; boosting; hierarchical models.

\section{Introduction}
\label{sec:intro}
Recent technological advances in biological measurement enable the collection of quasi-continuous temporal information on human health, including glucose concentrations, step counts, heart rate, and other relevant physiological and functional-capacity parameters \cite{matabuena2019application}. A common feature of such medical data is that multiple observations are collected from each individual, often on different time scales \cite{matabuena2022estimating}. For example, daily glucose-profile functions can be observed over multiple days for the same person \cite{matabuena2026beyond}.

In this setting, statistical and machine-learning methods for random-effects modeling provide an appropriate analytical framework for estimation, inference, and personalized prediction~\cite{gelman2007data,laird1982random}. This view is supported by an extensive theoretical and applied literature on random-effects models for scalar outcomes, with applications in areas such as economics, ecology, medicine, agriculture, and beyond. More recently, methodology for random-effects modeling in Hilbert spaces has emerged \cite{FDAwithR}, although much of this work has focused on linear functional models \cite{crainiceanu2012bootstrap,10.1093/biostatistics/kxs051,doi:10.1080/10618600.2021.1950006}. These developments have been motivated in part by the analysis of data from wearable and implantable technologies, such as accelerometers and other monitoring devices, where signals are often treated as raw or smoothed time series \cite{Greven2010}.

Continuing with the medical-science motivation example, quasi-continuous information is increasingly collected under real-world conditions. In such settings, biological signals cannot always be directly compared as synchronized time series \cite{matabuena2021glucodensities}. For example, individuals may exercise or consume meals at different times, making direct temporal alignment difficult or inappropriate. To address this issue and obtain meaningful summaries of such signals, some authors have proposed representing time series through their distributions \cite{matabuena2024multilevelfunctionaldataanalysis}, that is, as random objects taking values in a metric space. To the best of our knowledge, existing methods include random-effects models for univariate distributions \cite{matabuena2024multilevel} and related approaches for random objects on geodesic spaces \cite{bhattacharjee2023geodesic}, but these methods are typically linear and do not provide a directly comparable framework for repeated responses taking values in a general metric space. Thus, there is currently no standard random-effects baseline for the full setting considered here. Consequently, there remains a need for broadly applicable random-effects methodologies in metric spaces.

This paper introduces a random-effects algorithm for modeling arbitrary random objects defined on a metric space $(\Omega, d)$ \cite{frechet1948elements}, motivated by applications in biomedicine. The proposed method is computationally simple, readily parallelizable, and offers a flexible framework for capturing nonlinear statistical associations with metric-space-valued outcomes.


The core contributions of this paper are summarized in the following.

\subsection{Scientific contributions}

\begin{itemize}
    \item \textbf{A general regression framework for random effects in metric spaces.} 
    We propose a general strategy for developing novel regression algorithms for random effects in metric spaces. The approach extends massive univariate fitting strategies, commonly used in functional data analysis, to metric-space-valued data. The algorithm is easily parallelizable, easily integrated with subsampling-based methods, and flexible enough to accommodate a wide range of base learners, including boosting and random forests.
    \item \textbf{Theory.} 
    We establish consistency results for the fixed effects within the proposed framework using the theory of M-estimators.

    \item \textbf{Computation.} 
    We develop a practical implementation based on scalar nonlinear boosting algorithm, which offers a flexible trade-off between statistical accuracy and computational cost~\cite{grinsztajn2022treebased}.

    \item \textbf{Real-world performance.} 
    Extensive simulations and clinical case studies demonstrate the empirical performance of the proposed method and its versatility in handling various data structures that arise in modern clinical applications.
\end{itemize}

\subsection{Related metric-space literature}

Random-effects models for responses in general metric spaces remain relatively underdeveloped. Existing approaches include linear random-effects models for objects in geodesic spaces and for univariate probability distributions equipped with the \(2\)-Wasserstein metric. At the same time, there has been substantial recent progress on Fréchet-type methods for metric-space-valued data, including Wasserstein regression \cite{fan2021conditional,chen2021wasserstein,petersen2021wasserstein,zhou2021dynamic,dubey2022modeling,10.3150/21-BEJ1410,kurisumodel,chen2023sliced}, hypothesis testing \cite{10.1214/20-AOP1504,dubey2019frechet,petersen2021wasserstein,fout2023fr}, variable selection \cite{tucker2021variable}, missing-data methods \cite{Matabuena02102023}, multilevel modeling \cite{matabuena2024multilevel,bhattacharjee2023geodesic}, dimension reduction \cite{zhang2022nonlinear}, semiparametric regression \cite{bhattacharjee2021single,ghosal2023predicting}, semi-supervised learning \cite{qiu2024semisupervised}, and nonparametric regression \cite{schotz2021frechet,hanneke2022universally,bulte2023medoid,bhattacharjee2023nonlinear}. Nevertheless, these methods generally do not provide a directly comparable framework for repeated responses in arbitrary metric spaces with subject-specific random effects.

\subsection{Outline}

Section~\ref{sec:models} introduces the random-effects algorithm in metric spaces. Section~\ref{sec:real} illustrates the versatility of the approach in different data structures, including Laplacian graphs, probability distributions, and random functions in \(L^2\). Finally, Section~\ref{sec:conc} discusses the limitations of the work, summarizes the main findings, and suggests directions for future research. Formal proofs, a detailed description of the dataset, and simulation experiments based on synthetic data, designed to assess the finite-sample performance of the proposed method under controlled conditions, are presented in the Supplementary Material.

\section{Random-Effects Algorithm in Metric Spaces}\label{sec:models}

For scalar responses \(Y_{ij}\in\mathbb R\), where \(i=1,\dots,n\) indexes individuals and \(j=1,\dots,n_i\) indexes repeated visits, an additive Gaussian mixed model \cite{zuur2009mixed} is considered:
\[
Y_{ij}
=
\mu(X_{ij})+b_i+\varepsilon_{ij},
\quad
b_i \stackrel{\mathrm{iid}}{\sim} \mathcal N(0,\sigma_b^2),
\quad
\varepsilon_{ij} \stackrel{\mathrm{iid}}{\sim} \mathcal N(0,\sigma^2),
\quad
i=1,\dots,n,\quad j=1,\dots,n_i .
\]
\noindent Here \(b_i\) is independent of \(\varepsilon_{ij}\), \(\mu(\cdot)\) is a possibly nonlinear fixed-effect regression function, \(X_{ij}\in\mathcal X\) denotes fixed effect covariates, \(b_i\) is a Gaussian individual-specific random effect that captures dependence and heterogeneity within the subject, and \(\varepsilon_{ij}\) is a Gaussian observation-specific residual error. This additive formulation does not extend directly to responses in a general metric space \((\Omega,d)\) \cite{frechet1948elements}, since such spaces do not necessarily support addition, subtraction, or other linear operations. Examples include probability distributions, networks, covariance matrices, and other random objects that do not possess an intrinsic linear vector-space structure.

Let us define the observed data set:
\[
\mathcal D
=
\{(X_{ij},Y_{ij}): i=1,\dots,n,\ j=1,\dots,n_i\},
\qquad
X_{ij}\in\mathcal X,\quad Y_{ij}\in\Omega .
\]
\noindent
Assume that each individual has an unobserved random effect \(B_i\in\mathcal B\), where \(\mathcal B\) may be finite-dimensional or itself a structured metric space. In the metric-space setting, the random effect is not added to the response. Instead, it indexes the individual-specific conditional distribution of \(Y\). For \(x\in\mathcal X\), \(b\in\mathcal B\), and \(\omega\in\Omega\), define the conditional Fréchet risk of the oracle as
\[
M(\omega;x,b)
:=
\mathbb E\!\left\{d^2(Y,\omega)\mid X=x,\ B=b\right\}.
\]

\noindent Assuming the existence of a minimizer, the subject-specific Fréchet regression target of the oracle is
\[
m(x,b)
\in
\arg\min_{\omega\in\Omega} M(\omega;x,b).
\]
\noindent

\noindent This target is oracle because the random effect \(B\) is latent. In practice, \(B_i\) is not estimated directly as an element of the original space \(\mathcal B\). Instead, we estimate an anchor-based representation of the subject-specific random effect from the repeated observations available for individual \(i\).  Let
\[
I_n=\{(i,j): i=1,\dots,n,\ j=1,\dots,n_i\}
\]
denote the set of all observed subject--visit indices. Each anchor index
\(\ell=(i_\ell,j_\ell)\in I_n\) defines an observed reference object $
A_\ell=Y_{i_\ell j_\ell}\in\Omega$. The anchors are therefore observed responses that are used as fixed reference points in the metric space.

For each anchor \(A_\ell\), we transform the metric-space response into the scalar variable
\[
Z_{ij}^{(\ell)}
=
\log\!\left\{d^2(Y_{ij},A_\ell)+\delta\right\},
\qquad \delta>0,
\]
\noindent
where \(\delta\) is included for numerical stability. For each anchor, we fit the scalar mixed-effects model
\[
Z_{ij}^{(\ell)}
=
f_\ell(X_{ij})
+
b_i^{(\ell)}
+
\varepsilon_{ij}^{(\ell)} .
\]
\noindent
Here, \(b_i^{(\ell)}\) is an anchor-specific scalar random effect. It represents the contribution of subject \(i\)'s latent effect to the expected transformed distance between \(Y_{ij}\) and the anchor \(A_\ell\). Thus, \(B_i\) denotes the conceptual latent subject-specific random effect in the oracle model, whereas the collection of anchor-specific random effects $B_i^A
=
\{b_i^{(\ell)}:\ell\in I_n\}$ provides an anchor-based representation of this latent effect. This representation is empirical and anchor-dependent: it does not estimate \(B_i\) directly as an element of \(\mathcal B\), but summarizes its subject-specific contribution through transformed distances to the observed anchors.

For individual \(i\), the fitted model yields the subject-specific predicted transformed distance
\[
\widehat Z_i^{(\ell)}(x)
=
\widehat f_\ell(x)+\widehat b_i^{(\ell)},
\]
\noindent
where \(\widehat b_i^{(\ell)}\) is the estimated random effect for individual \(i\) in the \(\ell\)th anchor model. The estimated collection
$\widehat B_i^A
=
\{\widehat b_i^{(\ell)}:\ell\in I_n\}$ therefore serves as the empirical anchor-based representation of the latent random effect \(B_i\).

For individual \(i\), the final metric-space prediction is obtained by solving a discrete optimization problem on the observed response objects. Let $
\mathcal Y_n
=
\{Y_{ij}: i=1,\dots,n,\ j=1,\dots,n_i\}
\subset \Omega$ denote the empirical response space. We define
\[
\widehat m_{n,i}(x)
\in
\arg\min_{\omega\in \mathcal Y_n}
\sum_{\ell\in I_n}
\left[
\log\!\left\{d^2(\omega,A_\ell)+\delta\right\}
-
\left\{\widehat f_\ell(x)+\widehat b_i^{(\ell)}\right\}
\right]^2 .
\]
\noindent
Thus, the estimator selects the observed response object whose transformed distances to the anchors best match the individual-specific predicted transformed distances. The optimization is carried out over the finite empirical response space \(\mathcal Y_n\), while personalization enters through the estimated anchor-specific random effects \(\widehat b_i^{(\ell)}\). Because this criterion matches predicted and observed transformed anchor-distance profiles rather than directly minimizing the conditional Fréchet risk, the proposed method can be viewed as an anchor-based surrogate for subject-specific Fréchet regression in a general metric space.

\paragraph{Computational details.}
For each anchor \(\ell\in I_n\), we fit a scalar mixed-effects regression model to the transformed distances \(Z_{ij}^{(\ell)}\). Specifically, we use GPBoost~\cite{sigrist2022gpboost}, which combines gradient-boosted trees~\cite{prokhorenkova2018catboost} with Gaussian random-effect components~\cite{hajjem2014mixed}. For each fixed anchor \(\ell\), the working model is
\begin{equation}
Z_{ij}^{(\ell)}
=
f_\ell(X_{ij})
+
b_i^{(\ell)}
+
\varepsilon_{ij}^{(\ell)},
\qquad
b_i^{(\ell)}\stackrel{\mathrm{iid}}{\sim}\mathcal N(0,\sigma_{b,\ell}^2),
\qquad
\varepsilon_{ij}^{(\ell)}\stackrel{\mathrm{iid}}{\sim}\mathcal N(0,\sigma_\ell^2).
\label{eq:gpboost_model}
\end{equation}
Here, \(f_\ell:\mathcal X\to\mathbb R\) is a boosted-tree regression function, \(b_i^{(\ell)}\) is an anchor-specific individual random intercept, and \(\varepsilon_{ij}^{(\ell)}\) is an observation-level error. For each fixed anchor \(\ell\), the random intercepts \(b_i^{(\ell)}\), \(i=1,\dots,n\), are modeled as independent Gaussian subject effects. The anchor-specific models are fitted separately; therefore, possible dependence among \(b_i^{(\ell)}\) across different anchors is not modeled explicitly and is treated as part of the working-model approximation. Thus, each anchor defines one scalar mixed-effects regression problem, providing a natural extension of massively univariate mixed-effects modeling approaches used in the \(L^2\)-based functional data literature \cite{cui2021fast} to the metric-space setting. The boosted-tree component and variance parameters are estimated jointly, within each anchor-specific model, using GPBoost's hybrid boosting--mixed-effects procedure. In our implementation, we used 200 boosting iterations, a learning rate \(\eta=0.05\), a maximum tree depth of 6, 31 leaves, and a minimum of 5 observations per leaf. The pseudocode is provided in in the Supplemental Material. Although our empirical analyses focus on anchor-specific random intercepts, the proposed framework can accommodate richer mixed-effects structures, including random slopes, nested effects, crossed effects, or temporal random effects, by replacing the scalar anchor-level working model with the corresponding mixed-effects specification.

\subsection{Random-effect model theory}\label{sec:theory}

The goal of this section is to provide a theoretical characterization of our modeling strategy for metric-space-valued outcomes. In particular, we study conditions under which the proposed estimator remains consistent, even when the random-effect structure is used only as a working model.

In general, theory for multidimensional Euclidean and infinite-dimensional random-effects models remains relatively underdeveloped. Existing results for linear random-effects models often focus on asymptotic distributions of model parameters \cite{jiang2022usable}, while classical consistency results typically study fixed-effect estimation under possibly misspecified random-effect structures \cite{qu2000improving}
. Here, we extend this perspective to metric-space-valued outcomes. Conditional on $z=(x,b)\in\mathcal X\times\mathcal B$, and under suitable assumptions on the population and empirical M-estimation problem \cite{geer2000empirical}, we establish conditions under which

\[
d\bigl(\widehat m_n(z),m(z)\bigr)\xrightarrow{P}0.
\]
The supplementary material contains the formal proofs, together with extensions establishing uniform consistency.

\begin{proposition}[Pointwise consistency from criterion convergence]
Fix \(z=(x,b)\in \mathcal{X} \times \mathcal{B}\). Let \(\Omega_0\subset\Omega\) contain the true minimizer \(m(z)\), and suppose that \(\widehat m_n(z)\in\Omega_0\). Assume:
\begin{enumerate}
    \item \(m(z)\) is the unique minimizer of \(M(\cdot,z)\) over \(\Omega_0\);
    \item for every \(\varepsilon>0\),
    \[
    c_\varepsilon(z)
    :=
    \inf_{\omega\in\Omega_0:\ d(\omega,m(z))\ge \varepsilon}
    \{M(\omega,z)-M(m(z),z)\}
    >0;
    \]
    \item
    \[
    \sup_{\omega\in\Omega_0}
    \bigl|\widehat M_n(\omega,z)-M(\omega,z)\bigr|
    \xrightarrow{P}0;
    \]
    \item
    \[
    \widehat M_n(\widehat m_n(z),z)
    \le
    \inf_{\omega\in\Omega_0}\widehat M_n(\omega,z)+r_n(z),
    \qquad r_n(z)\xrightarrow{P}0.
    \]
\end{enumerate}
Then
\[
d\bigl(\widehat m_n(z),m(z)\bigr)\xrightarrow{P}0.
\]
\end{proposition}

\paragraph{Interpretation of the assumptions.}
Assumption (1) defines a unique target \(m(z)\). Assumption (2) imposes separation, ensuring that points away from \(m(z)\) have a strictly larger risk. Assumption (3) requires uniform convergence of the empirical criterion to the population criterion over \(\Omega_0\). Assumption (4) allows for an approximate numerical minimization, so \(\widehat m_n(z)\) needs only be an asymptotic minimizer of \(\widehat M_n(\cdot,z)\).

In the Supplemental Material, we provide sufficient conditions for the anchor-based GPBoost implementation under which the consistency results hold, including correct specification or consistent estimation of the scalar anchor-level working models.

\section{Case Studies in Digital Health Cohorts}
\label{sec:real}

The main goal of this section is to demonstrate the versatility of the proposed random-effects algorithm for metric-space-valued responses that arise in modern digital health applications. We consider several types of random objects as responses: (i) random functions in Hilbert space $L^{2}$, (ii) probability distributions equipped with the Wasserstein metric $\mathcal{W}_{2}$, and (iii) Laplacian graphs equipped with the Frobenius metric. Our specific aims are twofold: (i) to assess whether incorporating individual random effects improves predictive performance relative to a fixed-effects-only model, and (ii) to highlight the practical relevance of the proposed framework for analyzing high-frequency clinical data. In Supplemental Material, we complement this analysis with finite-sample performance studies in settings where the ground-truth comparison benchmark is known or can be approximated using Monte Carlo methods.

Digital health technologies generate unprecedented volumes of high-frequency longitudinal data on physiology, behavior, and environmental exposures. Wearable devices, such as continuous glucose monitors (CGMs), provide detailed real-world information on metabolic control, disease progression, and response to treatment \cite{doi:10.1056/NEJMoa0805017,matabuena2021glucodensities}. However, most current methods reduce these rich signals to a small number of summary measures, thus overlooking their functional, distributional, or temporal dynamic structure and limiting the clinical knowledge that can be extracted \cite{matabuena2024multilevelfunctionaldataanalysis,Matabuena2025}. This creates a methodological gap between the complexity of the data currently being collected, the practical opportunities these data offer, and the analytical tools available to translate them into robust clinical and population-level evidence.

We apply our method to several metric-space representations of dense time series collected in digital health studies. In total, we consider four case studies: two based on continuous glucose monitoring data from CGMCR and two based on physical activity data from NHANES study. Together, these studies illustrate distinct ways of representing observed time series as metric-space-valued objects.

In all cases, performance evaluation is performed at the individual level to mimic the goal of personalized prediction. For each individual $i$, repeated observations are divided into training and holdout subsets. Formally, we partition the index set as \(I_n = I_{\mathrm{train}}\cup I_{\mathrm{test}}\), with \(I_{\mathrm{train}}\cap I_{\mathrm{test}}=\varnothing\), and write \(I_{\mathrm{train}}^{i}=\{j:(i,j)\in I_{\mathrm{train}}\}\) and \(I_{\mathrm{test}}^{i}=\{j:(i,j)\in I_{\mathrm{test}}\}\) for the replicate indices of individual \(i\) in the training and holdout splits, respectively. Training replicas are used to estimate individual-specific random-effect components, and only training responses \(\{Y_\ell : \ell \in I_{\mathrm{train}}\}\) are used as candidate anchors; held-out responses are never used as anchors. This design allows us to test whether information from previous observations of the same individual improves prediction on future or held-out replicates. Unless otherwise stated, we use \(60\%\) of each individual's observations for training and the remaining \(40\%\) for formal test evaluation.

We evaluate prediction accuracy using the mean squared distance. For a prediction strategy
\(s\in\{\mathrm{RE},\mathrm{FE}\}\), the individual-level prediction error is
\begin{equation}
    \mathrm{MSE}_i^{(s)}
    =
    \frac{1}{|I_{\mathrm{test}}^{i}|}
    \sum_{j \in I_{\mathrm{test}}^{i}}
    d^{2}\!\left(Y_{ij},\widehat{m}_n^{(s)}(X_{ij},i)\right),
    \qquad i=1,\dots,n.
    \label{eq:frechet_mse_real}
\end{equation}
\noindent Here, \(\widehat{m}_n^{\mathrm{RE}}\) incorporates the estimated individual-specific random-effect component, whereas \(\widehat{m}_n^{\mathrm{FE}}\) denotes the fixed-effects-only predictor and does not use individual-specific random effects.

\subsection{\texorpdfstring{$L^{2}$}{L2} Hilbert-space benchmarks: NHANES and CGMCR datasets}
\label{sec:real_euclidean}

\begin{figure*}[t]
    \centering
    \includegraphics[width=0.9\textwidth]{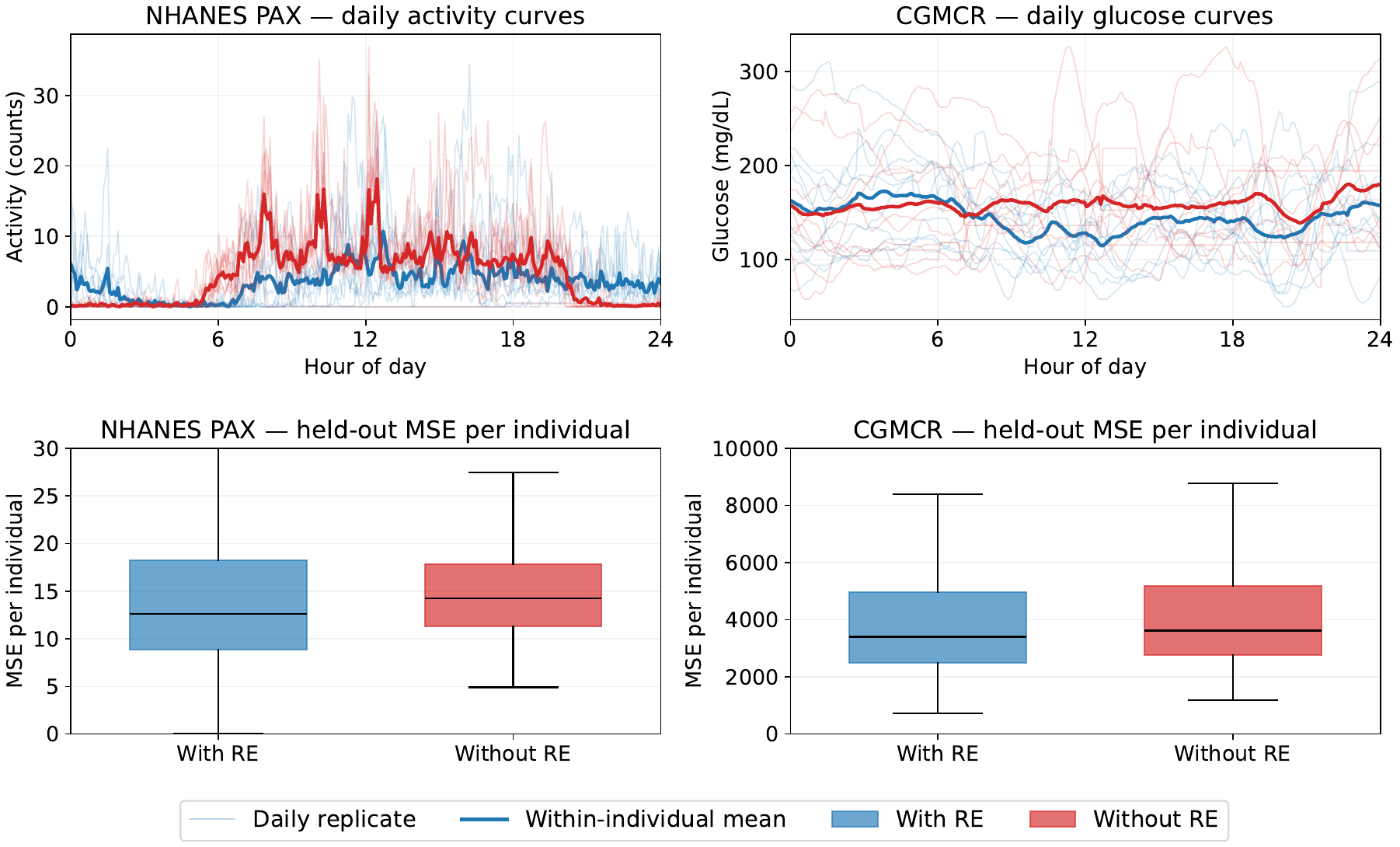}
    \caption{$L^{2}$ Hilbert-space examples for physical activity and continuous glucose monitoring data. \textbf{Top row:} representative repeated daily curves for NHANES PAX (left) and CGMCR (right). Faint lines denote daily replicates, and solid lines denote within-individual means for two representative individuals in each dataset. \textbf{Bottom row:} held-out per-individual MSE for \emph{With RE} and \emph{Without RE} on NHANES PAX (left) and CGMCR (right).}
    \label{fig:real_euclidean_curves}
\end{figure*}

We first consider two functional-response datasets in which each individual observation is a daily trajectory in Hilbert space \(L^{2}([0,24])\): accelerometry-derived physical-activity data from NHANES and continuous glucose-monitoring data from CGMCR. This setting is natural because functional mixed-effects models are well-established in Hilbert spaces, whereas our anchor-based formulation uses the trajectories only through pairwise distances. Each candidate anchor is a training daily curve, and the prediction selects the anchor minimizing the fit conditional squared distance \(L^{2}\).

For two daily trajectories \(Y_{ij}\) and \(Y_{i'j'}\), the loss is based on the squared \(L^{2}\) distance,
\begin{equation}
    d^{2}(Y_{ij},Y_{i'j'})
    =
    \|Y_{ij}-Y_{i'j'}\|^{2}_{L^{2}}
    =
    \int_{0}^{24}
    \{Y_{ij}(t)-Y_{i'j'}(t)\}^{2}\,dt .
    \label{eq:l2_distance_real}
\end{equation}

The NHANES dataset contains minute-level physical-activity counts derived from accelerometry over 24 hours. After filtering for complete covariates and restricting to individuals younger than \(40\) years, \(n=3{,}727\) individuals remain, each contributing up to 10 daily curves. Sex, age, BMI, and waist circumference are used as predictors.

The CGMCR dataset contains continuous glucose-monitoring data from individuals with Type~1 diabetes, sampled every five minutes over 24 hours. After filtering for complete covariates and requiring at least six valid days, \(n=344\) individuals remain, each contributing up to 40 daily curves. Age and sex are used as fixed-effect predictors. In both datasets, we consider only random intercepts. Repeated daily curves from the same individual are divided into training and held-out subsets, following the protocol described above. Training curves are used as candidate anchors and as individual-specific random intercepts in anchor-distance regressions. 

Figure~\ref{fig:real_euclidean_curves} illustrates the data structure for two representative individuals from each data set. The upper row shows repeated daily curves, with faint lines denoting daily replicates and solid lines denoting within-individual means. The glucose curves exhibit stronger heterogeneity between-individuals, whereas the physical-activity curves appear more regular throughout the days. The lower row reports the individual MSE held for anchor-based predictors with and without random effects.

Both datasets show consistent gains from incorporating individual-specific random effects. In NHANES, the holdout MSE decreases from \(16.26\) to \(14.64\), a relative reduction of about \(10.0\%\). In CGMCR, the MSE decreases from \(4{,}368\) to \(3{,}994\), corresponding to an \(8.6\%\) reduction. These improvements suggest that the proposed anchor-based formulation effectively leverages repeated observations to capture individual-specific variation and improve personalized prediction. Compared with classical functional linear random-effects models \cite{cui2021fast}, our nonlinear random-effects approach achieves lower MSE in both CGMCR (\(3{,}994\) vs. \(4{,}013.91\)) and NHANES (\(14.64\) vs. \(15.00\)), indicating their superiority to capture stronger nonlinear statistical associations in these datasets.

\subsection{CGM 3D glucose distributions}
\label{sec:cgm_dist}

We again consider the CGMCR dataset, but now represent each individual through block-level distributions of glucose dynamics rather than daily glucose curves. We use a 40-day observation period for each individual and include \(n=415\) individuals. We use the same set of fixed-effect covariates as above. Because the preprocessing differs from the functional-curve analysis, this sample includes a larger number of participants.

For each individual, the 40-day observation period is divided into four non-overlapping 10-day blocks. Within each block, and for each available time point \(t \in \mathcal{T}\), we consider the glucose level \(G(t)\), glucose velocity \(\Delta G(t)/\Delta t\), and glucose acceleration \(\Delta^2 G(t)/\Delta t^2\). Following the distributional representation introduced in \cite{Matabuena2025}, these three quantities are used to construct a multivariate distributional summary of glucose dynamics for each block. Thus, the clinical outcome \(Y_{ij}\), \(j=1,\dots,4\), is represented as a probability distribution in \(\mathbb{R}^3\). We endow this probability distribution space with the 2-Wasserstein metric. The squared 2-Wasserstein metric between two probability measures \(\mu\) and \(\nu\) with finite second moments is defined as
\begin{equation}
\mathcal{W}_2^2(\mu,\nu)
=
\inf_{\pi \in \Pi(\mu,\nu)}
\int_{\mathbb{R}^d \times \mathbb{R}^d}
\|u-v\|^2 \, d\pi(u,v),
\end{equation}
\noindent where \(\Pi(\mu,\nu)\) denotes the set of couplings of \(\mu\) and \(\nu\), namely the set of probability measures on \(\mathbb{R}^d \times \mathbb{R}^d\) with marginals \(\mu\) and \(\nu\). This metric compares two distributions by finding the least costly way to transport the probability mass from one distribution to the other. 

\begin{figure}[t]
    \centering
    \includegraphics[width=0.9\columnwidth]{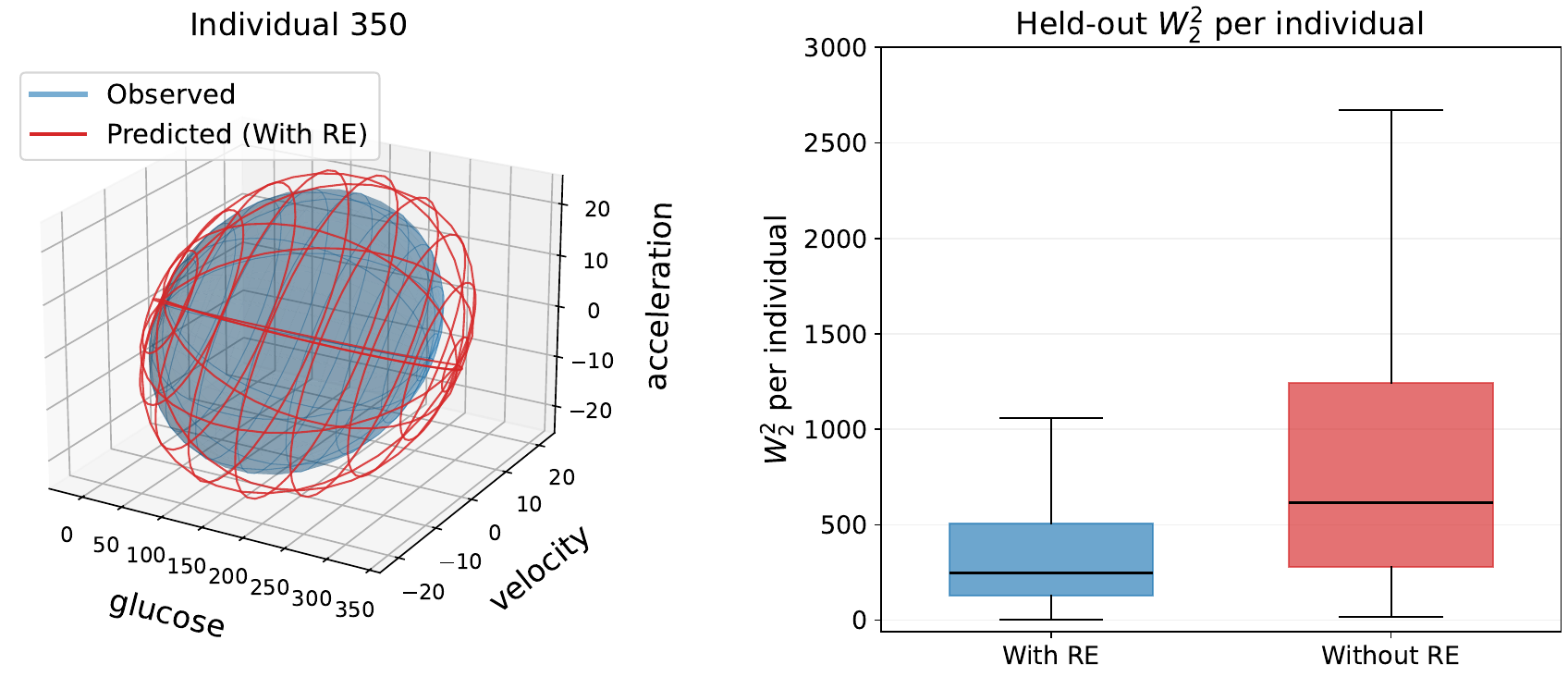}
    \caption{CGM 3D distributions. \textbf{Left}: observed \(95\%\) probability ellipsoid of the distributional response for one held-out individual (blue, filled), together with the ellipsoid of the anchor selected by \emph{With RE} (red, wire); axes are in the native CGM-derived units. \textbf{Right}: held-out per-individual $W_2^2$ error under \emph{With RE} and \emph{Without RE}, summarizing the distribution of individual-level errors across the \(415\) individuals.}
    \label{fig:real_results_cgm_distributions}
\end{figure}

Figure~\ref{fig:real_results_cgm_distributions} shows the practical advantage of incorporating random effects to reduce the prediction error. The figure also illustrates the structure of the distributional prediction for the individual 350. The \emph{With RE} strategy achieves a mean waterstein squared error of \(421\), whereas the \emph{Without RE} strategy has an error of approximately \(986\), corresponding to a reduction of \(57\%\) relative to the fixed-effects only strategy. In this dataset, the random effect is not just a small correction to the population-level predictor but rather the main mechanism that allows for personalized prediction.

The proposed glucose distributional representation captures not only average glucose levels and distributional features, such as quantiles, but also local glucose-regulation dynamics through the inclusion of glucose velocity and acceleration in the joint distribution. The multivariate distributional example highlights a setting in which the proposed framework and standard additive random-effects formulations are not directly applicable. Although there are related methods for univariate probability distributions, the multivariate setting considered here requires a more general framework.

\subsection{Laplacian graph example in NHANES}
\label{sec:nhanes_graphs}

\begin{figure*}[t]
    \centering
    \includegraphics[width=\textwidth]{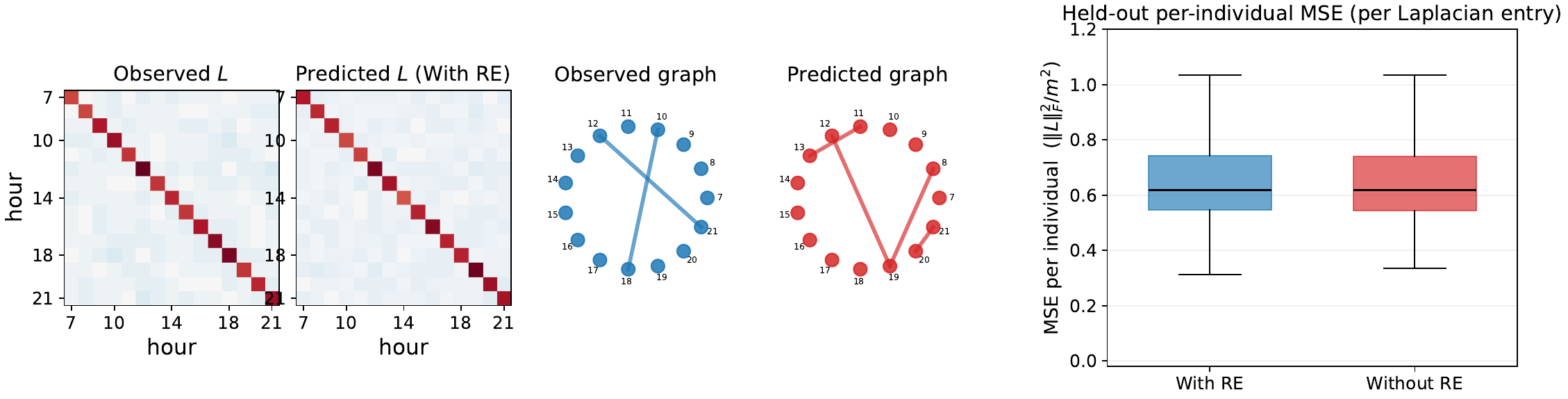}
    \caption{NHANES hourly correlation graphs. \textbf{From left to right}: observed Laplacian and corresponding thresholded graph for one held-out individual-day pair; Laplacian and graph selected by the \emph{With RE} strategy for the same case; and held-out per-individual squared Frobenius error under \emph{With RE} and \emph{Without RE}.}
    \label{fig:real_nhanes_graphs_combined}
\end{figure*}

Finally, we consider a graph-derived response constructed from the NHANES accelerometry data. Instead of representing each day by its activity trajectory, we represent each individual-day pair by the temporal dependence structure of activity across daytime hours.

Let \(\mathcal H=\{7,\ldots,21\}\) denote the daytime hours used to define the outcome. For each individual-day pair and each hour \(h\in\mathcal H\), we collect the twelve 5-minute activity readings into a vector
\[
    \mathbf a_h \in \mathbb R^{12}.
\]
We then compute the \(H\times H\) Pearson correlation matrix
\[
    C_{hh'}
    =
    \operatorname{corr}(\mathbf a_h,\mathbf a_{h'}),
    \qquad h,h'\in\mathcal H.
\]
\noindent Individual-day pairs with undefined correlation entries are discarded. A binary adjacency matrix $A$ is obtained by thresholding the correlations such that $A_{hh'}=1$ if $|C_{hh'}|>\tau$ and $A_{hh'}=0$ otherwise, with $\tau=0.3$. Diagonal entries are set to zero, $A_{hh}=0$, to exclude self-connections. The corresponding graph Laplacian is
\[
    L = D-A,
    \qquad
    D=\operatorname{diag}(A\mathbf 1).
\]
Each daily Laplacian is treated as one replication. Distances between graph-valued responses are measured using the squared Frobenius distance \cite{zhou2022network}, normalized by the number of Laplacian entries:
\begin{equation}
d^2(L,L')
=
\frac{1}{H^2}\|L-L'\|_F^2.
\end{equation}

This representation captures the temporal correlation structure of daily activity rather than its overall magnitude. Unlike the $L^2$ setting, in which each response is a time trajectory, the graph response summarizes how activity profiles covary across hours of the day. Because this response reflects second-order temporal dependence, we use an extended set of 10 predictors: sex, age, BMI, waist circumference, systolic and diastolic blood pressure, grip strength, gamma-glutamyl transferase, triglycerides, and creatinine. After excluding individuals with missing covariates and restricting the sample to those younger than $40$ years, $n=2{,}091$ individuals remain, each contributing up to 10 daily Laplacians split into training and evaluation subsets. Additional processing details are provided in the Supplemental Material.

In Figure~\ref{fig:real_nhanes_graphs_combined}, the \emph{With RE} and \emph{Without RE} strategies yield nearly identical held-out errors, with MSE values of $0.664$ and $0.665$, respectively. Thus, in this setting, incorporating subject-specific random effects provides little additional predictive improvement. This result is nevertheless informative. The thresholded graph construction is sensitive to local changes in hourly correlations, so the resulting Laplacians can vary substantially across days for the same individual. Consequently, although subject-level heterogeneity may be present, the repeated observations may not contain a sufficiently stable individual-specific signal to improve prediction beyond the fixed-effects-only model.  Future work could explore alternative graph constructions or prediction models that better capture stable subject-specific dependence.

\subsection{Summary of results}

\begin{figure*}[t]
\centering

\begin{minipage}[t]{0.47\textwidth}
\centering
\textbf{(a) Summary of real-data results}

\vspace{0.15cm}

\scriptsize
\setlength{\tabcolsep}{2.5pt}
\renewcommand{\arraystretch}{0.85}
\resizebox{\textwidth}{!}{%
\begin{tabular}{lcccc}
\toprule
\textbf{Dataset} & \textbf{Metric} & \textbf{$n$} & \textbf{With RE} & \textbf{Without RE}  \\
\midrule
CGMCR             & $\|\cdot\|_{L^2}^2$          & 344     & 3{,}994 & 4{,}368  \\
NHANES PAX        & $\|\cdot\|_{L^2}^2$          & 3{,}727 & 14.64   & 16.26    \\
CGM distributions & $W_2^2$                      & 415     & 421     & 985.6    \\
NHANES graphs     & $\frac{1}{H^2}\|\cdot\|_F^2$ & 2{,}091 & 0.664   & 0.665   \\
\bottomrule
\end{tabular}%
}
\end{minipage}
\hfill
\begin{minipage}[t]{0.50\textwidth}
\centering
\textbf{(b) Per-individual held-out errors}

\vspace{0.15cm}

\includegraphics[width=\textwidth]{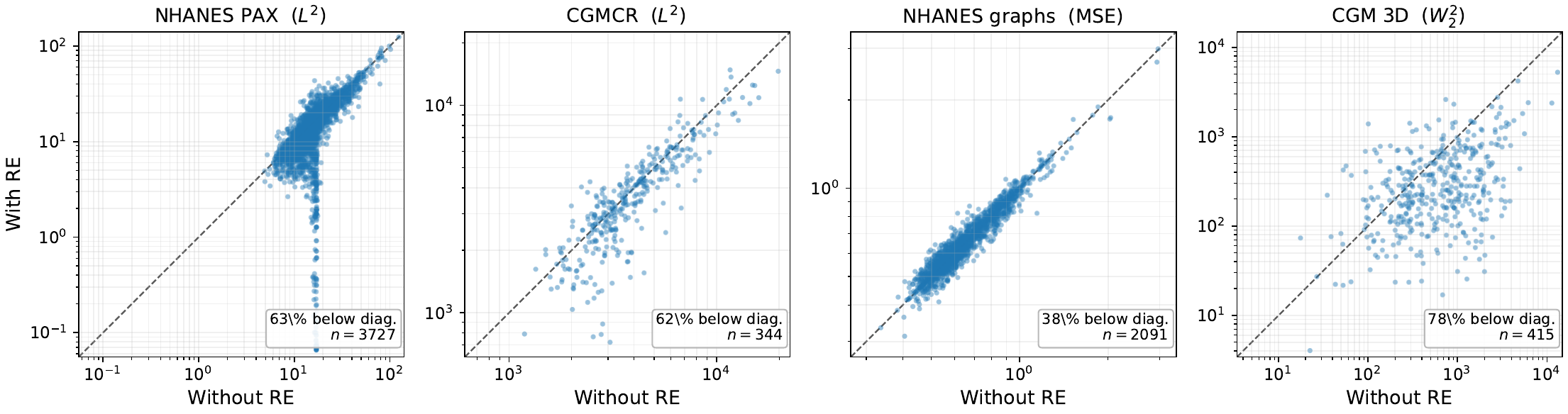}
\end{minipage}

\vspace{0.2cm}

\caption{Summary of real-data prediction results across the four datasets analyzed. Panel (a) reports the mean held-out squared distance for the \emph{With RE} and \emph{Without RE} predictors in the natural metric of each response space. Panel (b) shows the corresponding per-individual held-out errors, with \emph{With RE} on the \(y\)-axis and \emph{Without RE} on the \(x\)-axis. Axes are shown on a log--log scale, and the dashed diagonal denotes equal performance. Points below the diagonal correspond to individuals for whom the random-effects predictor improves held-out prediction.}
\label{fig:real_summary_combined}

\end{figure*}

Table~\ref{fig:real_summary_combined} summarizes performance across the four real-data benchmarks. Random effects improve held-out prediction in three datasets. The greatest gain occurs for the CGM 3D distributions, where the mean squared Wasserstein error is reduced by about \(57\%\). The two \(L^2\) functional benchmarks show smaller but consistent improvements. In contrast, the NHANES hourly correlation graphs remain a near-null case, suggesting that subject-specific structure in the thresholded Laplacians is weak, unstable across days, or not well captured by the available data.

Figure~\ref{fig:real_summary_combined} shows the corresponding individual-level comparison. Errors without random effects are shown on the \(x\)-axis and errors with random effects on the \(y\)-axis; points below the diagonal indicate an improvement over random effects. The figure agrees with the aggregate results: CGM 3D distributions show the strongest improvement, with \(78\%\) of individuals below the diagonal, followed by NHANES and CGMCR, with \(63\%\) and \(62\%\), respectively. For NHANES graphs, only \(38\%\) of the individuals fall below the diagonal, indicating that there is no systematic benefit from random effects. The NHANES panel also reveals a small subgroup of \(83\) individuals, about \(2\%\) of the cohort, for whom \emph{With RE} nearly eliminates the prediction error while \emph{Without RE} remains much larger. These individuals do not differ markedly in the available demographic covariates, but their activity profiles are nearly flat, with mean activity counts per slot approximately \(20\times\) lower than the average of the cohort. Thus, the random effect is especially informative among highly sedentary individuals, indicating that personalized prediction is particularly effective for this subgroup.

\section{Final Remarks}\label{sec:conc}

This paper proposes a new class of random-effects models for random objects taking values in metric spaces \cite{frechet1948elements}. Its main limitation is computational, as the approach requires fitting multiple univariate random-effects models. However, for applications involving approximately $n=4000$ individuals and $n_i=10$ repeated measurements per individual, the proposed boosting-based algorithm computes the solution in less than one hour on a high-performance computing cluster.

Future work will focus on improving scalability through subsampling algorithms for large-scale applications. Another important direction is uncertainty quantification for responses valued in the metric-space \cite{politis1994large}. However, a full treatment of this topic, including conformal prediction \cite{lugosi2025conformal}, is beyond the scope of the present work and constitutes a substantial independent research direction. These problems are challenging and timely, especially in modern digital health settings, where repeated measurements naturally generate complex random objects in metric spaces and the ultimate analytical goal is personalized prediction.



\newpage

\printbibliography

@book{gelman2007data,
  title={Data analysis using regression and multilevel/hierarchical models},
  author={Gelman, Andrew and Hill, Jennifer},
  year={2007},
  publisher={Cambridge university press}
}

@inproceedings{prokhorenkova2018catboost,
  title     = {CatBoost: unbiased boosting with categorical features},
  author    = {Prokhorenkova, Liudmila and Gusev, Gleb and Vorobev, Aleksandr and Dorogush, Anna Veronika and Gulin, Andrey},
  booktitle = {Advances in Neural Information Processing Systems 31},
  pages     = {6639--6649},
  year      = {2018},
  doi       = {10.5555/3327757.3327770},
  url       = {https://papers.nips.cc/paper/7898-catboost-unbiased-boosting-with-categorical-features}
}

@article{grinsztajn2022treebased,
  title   = {Why do tree-based models still outperform deep learning on tabular data?},
  author  = {Grinsztajn, L{\'e}o and Oyallon, Edouard and Varoquaux, Ga{\"e}l},
  journal = {arXiv preprint arXiv:2207.08815},
  year    = {2022},
  doi     = {10.48550/arXiv.2207.08815},
  url     = {https://doi.org/10.48550/arXiv.2207.08815}
}

@article{sigrist2022gpboost,
  title   = {Gaussian Process Boosting},
  author  = {Sigrist, Fabio},
  journal = {Journal of Machine Learning Research},
  volume  = {23},
  number  = {232},
  pages   = {1--46},
  year    = {2022},
  url     = {http://jmlr.org/papers/v23/20-322.html}
}

@article{hajjem2014mixed,
  title   = {Mixed-effects random forest for clustered data},
  author  = {Hajjem, Ahlem and Bellavance, Fran{\c{c}}ois and Larocque, Denis},
  journal = {Journal of Statistical Computation and Simulation},
  volume  = {84},
  number  = {6},
  pages   = {1313--1328},
  year    = {2014},
  doi     = {10.1080/00949655.2012.741599}
}

@misc{matabuena2024multilevelfunctionaldataanalysis,
      title={Multilevel functional data analysis modeling of human glucose response to meal intake}, 
      author={Marcos Matabuena and Joe Sartini and Francisco Gude},
      year={2024},
      eprint={2405.14690},
      archivePrefix={arXiv},
      primaryClass={q-bio.QM},
      url={https://arxiv.org/abs/2405.14690}, 
}

@book{FDAwithR,
  title={{Functional Data Analysis with R}},
  author={Crainiceanu, C.M. and Goldsmith, J. and Leroux, A. and Cui, E.},
  year={2024},
  publisher={Chapman and Hall/CRC}
}

@article{ghosal2023predicting,
  title={Predicting distributional profiles of physical activity in the NHANES database using a Partially Linear Single-Index Fr$\backslash$'echet Regression model},
  author={Ghosal, Aritra and Matabuena, Marcos and Meiring, Wendy and Petersen, Alexander},
  journal={arXiv preprint arXiv:2302.07692},
  year={2023}
}

@article{doi:10.1080/10618600.2021.1950006,
author = {Erjia Cui and Andrew Leroux and Ekaterina Smirnova and Ciprian M. Crainiceanu},
title = {Fast Univariate Inference for Longitudinal Functional Models},
journal = {Journal of Computational and Graphical Statistics},
volume = {0},
number = {0},
pages = {1-12},
year  = {2021},
publisher = {Taylor & Francis},
doi = {10.1080/10618600.2021.1950006},

URL = { 
        https://doi.org/10.1080/10618600.2021.1950006
    
},
eprint = { 
        https://doi.org/10.1080/10618600.2021.1950006
    
}

}

@article{matabuena2019application,
  title={Application of functional data analysis for the prediction of maximum heart rate},
  author={Matabuena, Marcos and Vidal, Juan C and Hayes, Philip R and Saavedra-Garc{\'\i}a, Miguel and Trillo, Fernando Huelin},
  journal={IEEE Access},
  volume={7},
  pages={121841--121852},
  year={2019},
  publisher={IEEE}
}

@article{dubey2019frechet,
  title={Fr{\'e}chet analysis of variance for random objects},
  author={Dubey, Paromita and M{\"u}ller, Hans-Georg},
  journal={Biometrika},
  volume={106},
  number={4},
  pages={803--821},
  year={2019},
  publisher={Oxford University Press}
}

@article{doi:10.1056/NEJMoa0805017,
title = {Continuous Glucose Monitoring and Intensive Treatment of Type 1 Diabetes},
author={W.V. Tamborlane and R.W. Beck and B.W. Bode and others},
journal = {New England Journal of Medicine},
volume = {359},
number = {14},
pages = {1464-1476},
year = {2008}
}

@article{Matabuena2025,
  title        = {Glucodensity functional profiles outperform traditional continuous glucose monitoring metrics},
  author       = {Matabuena, Marcos and Ghosal, Rahul and Aguilar, Javier Enrique and Keshet, Ayya and Wagner, Robert and Fern{\'a}ndez Merino, Carmen and S{\'a}nchez Castro, Juan and Zipunnikov, Vadim and Onnela, Jukka-Pekka and Gude, Francisco},
  journal      = {Scientific Reports},
  year         = {2025},
  date         = {2025-09-29},
  volume       = {15},
  number       = {1},
  pages        = {33662},
  issn         = {2045-2322},
  doi          = {10.1038/s41598-025-18119-2},
  url          = {https://doi.org/10.1038/s41598-025-18119-2},
  note         = {Article number: 33662}
}

@Article{Greven2010,
author={Greven, Sonja
and Crainiceanu, Ciprian
and Caffo, Brian
and Reich, Daniel},
title={Longitudinal functional principal component analysis},
journal={Electronic journal of statistics},
year={2010},
volume={4},
pages={1022-1054},
note={21743825[pmid]},

issn={1935-7524},
doi={10.1214/10-EJS575},
url={https://pubmed.ncbi.nlm.nih.gov/21743825},

language={eng}
}

@article{cui2021fast,
  title={Fast univariate inference for longitudinal functional models},
  author={Cui, Erjia and Leroux, Andrew and Smirnova, Ekaterina and Crainiceanu, Ciprian M},
  journal={Journal of Computational and Graphical Statistics},
  pages={1--12},
  year={2021},
  publisher={Taylor \& Francis}
}

@article{10.1093/biostatistics/kxs051,
    author = {Gertheiss, Jan and Goldsmith, Jeff and Crainiceanu, Ciprian and Greven, Sonja},
    title = "{Longitudinal scalar-on-functions regression with application to tractography data}",
    journal = {Biostatistics},
    volume = {14},
    number = {3},
    pages = {447-461},
    year = {2013},
    month = {01},
    issn = {1465-4644},
    doi = {10.1093/biostatistics/kxs051},
    url = {https://doi.org/10.1093/biostatistics/kxs051},
    eprint = {https://academic.oup.com/biostatistics/article-pdf/14/3/447/17738955/kxs051.pdf},
}

@article{crainiceanu2012bootstrap,
  title={Bootstrap-based inference on the difference in the means of two correlated functional processes},
  author={Crainiceanu, Ciprian M and Staicu, Ana-Maria and Ray, Shubankar and Punjabi, Naresh},
  journal={Statistics in medicine},
  volume={31},
  number={26},
  pages={3223--3240},
  year={2012},
  publisher={Wiley Online Library}
}

@article{matabuena2022estimating,
  title={Estimating Knee Movement Patterns of Recreational Runners Across Training Sessions Using Multilevel Functional Regression Models},
  author={Matabuena, Marcos and Karas, Marta and Riazati, Sherveen and Caplan, Nick and Hayes, Philip R},
  journal={The American Statistician},
  number={just-accepted},
  pages={1--24},
  year={2022},
  publisher={Taylor \& Francis}
}

@article{petersen2021wasserstein,
  title={Wasserstein $ F $-tests and confidence bands for the Fr{\'e}chet regression of density response curves},
  author={Petersen, Alexander and Liu, Xi and Divani, Afshin A},
  journal={The Annals of Statistics},
  volume={49},
  number={1},
  pages={590--611},
  year={2021},
  publisher={Institute of Mathematical Statistics}
}

@article{matabuena2021glucodensities,
  title={Glucodensities: a new representation of glucose profiles using distributional data analysis},
  author={Matabuena, Marcos and Petersen, Alexander and Vidal, Juan C and Gude, Francisco},
  journal={Statistical Methods in Medical Research},
  volume={30},
  number={6},
  pages={1445--1464},
  year={2021},
  publisher={SAGE Publications Sage UK: London, England}
}

@article{politis1994large,
  title={Large sample confidence regions based on subsamples under minimal assumptions},
  author={Politis, Dimitris N and Romano, Joseph P},
  journal={The Annals of Statistics},
  pages={2031--2050},
  year={1994},
  publisher={JSTOR}
}

@article{lugosi2025conformal,
  title={Conformal and knn predictive uncertainty quantification algorithms in metric spaces},
  author={Lugosi, G{\'a}bor and Matabuena, Marcos},
  journal={arXiv preprint arXiv:2507.15741},
  year={2025}
}

@article{fan2021conditional,
  author = {Jianing Fan and Hans-Georg M{\"u}ller},
  title = {Conditional Distribution Regression for Functional Responses},
  journal = {Scandinavian Journal of Statistics},
  year = {2022},
  volume = {49},
  number = {2},
  pages = {502--524},
  doi = {10.1111/sjos.12525}
}

@article{chen2021wasserstein,
  author = {Yaqing Chen and Zhenhua Lin and Hans-Georg M{\"u}ller},
  title = {Wasserstein Regression},
  journal = {Journal of the American Statistical Association},
  year = {2023},
  volume = {118},
  number = {542},
  pages = {869--882},
  doi = {10.1080/01621459.2021.1956937}
}

@article{zhou2021dynamic,
  author = {Yidong Zhou and Hans-Georg M{\"u}ller},
  title = {Dynamic Network Regression},
  journal = {arXiv preprint arXiv:2109.02981},
  year = {2021},
  doi = {10.48550/arXiv.2109.02981}
}

@article{dubey2022modeling,
  author = {Paromita Dubey and Hans-Georg M{\"u}ller},
  title = {Modeling Time-Varying Random Objects and Dynamic Networks},
  journal = {Journal of the American Statistical Association},
  year = {2022},
  volume = {117},
  number = {540},
  pages = {2252--2267},
  doi = {10.1080/01621459.2021.1917416}
}

@article{10.3150/21-BEJ1410,
  author = {Jeong Min Jeon and Young Kyung Lee and Enno Mammen and Byeong U. Park},
  title = {Locally Polynomial Hilbertian Additive Regression},
  journal = {Bernoulli},
  year = {2022},
  volume = {28},
  number = {3},
  pages = {2034--2066},
  doi = {10.3150/21-BEJ1410}
}

@article{kurisumodel,
  author = {Daisuke Kurisu and Taisuke Otsu},
  title = {Model Averaging for Global Fr{\'e}chet Regression},
  journal = {Journal of Multivariate Analysis},
  year = {2025},
  volume = {207},
  pages = {105416},
  doi = {10.1016/j.jmva.2025.105416}
}

@article{chen2023sliced,
  author = {Han Chen and Hans-Georg M{\"u}ller},
  title = {Sliced Wasserstein Regression},
  journal = {arXiv preprint arXiv:2306.10601},
  year = {2023},
  doi = {10.48550/arXiv.2306.10601}
}

@article{10.1214/20-AOP1504,
  author = {Russell Lyons},
  title = {Second Errata to ``Distance Covariance in Metric Spaces''},
  journal = {The Annals of Probability},
  year = {2021},
  volume = {49},
  number = {5},
  pages = {2668--2670},
  doi = {10.1214/20-AOP1504}
}

@article{fout2023fr,
  author = {Alex Fout and Bailey K. Fosdick},
  title = {Fr{\'e}chet Covariance and MANOVA Tests for Random Objects in Multiple Metric Spaces},
  journal = {arXiv preprint arXiv:2306.12066},
  year = {2023},
  doi = {10.48550/arXiv.2306.12066}
}

@article{tucker2021variable,
  author = {Danielle C. Tucker and Yichao Wu and Hans-Georg M{\"u}ller},
  title = {Variable Selection for Global Fr{\'e}chet Regression},
  journal = {Journal of the American Statistical Association},
  year = {2023},
  volume = {118},
  pages = {1023--1037},
  doi = {10.1080/01621459.2021.1969240}
}

@article{Matabuena02102023,
  author = {Marcos Matabuena and Carla D{\'i}az-Louzao and Rahul Ghosal and Francisco Gude},
  title = {Personalized Imputation in Metric Spaces via Conformal Prediction: Applications in Predicting Diabetes Development with Continuous Glucose Monitoring Information},
  journal = {arXiv preprint arXiv:2403.18069},
  year = {2024},
  doi = {10.48550/arXiv.2403.18069}
}

@article{matabuena2024multilevel,
  title={Multilevel functional distributional models with applications to continuous glucose monitoring in diabetes clinical trials},
  author={Matabuena, Marcos and Crainiceanu, Ciprian M},
  journal={The Annals of Applied Statistics},
  volume={20},
  number={1},
  pages={476--495},
  year={2026},
  publisher={Institute of Mathematical Statistics}
}

@article{zhou2022network,
  title={Network regression with graph Laplacians},
  author={Zhou, Yidong and M{\"u}ller, Hans-Georg},
  journal={Journal of Machine Learning Research},
  volume={23},
  number={320},
  pages={1--41},
  year={2022}
}

@book{geer2000empirical,
  title={Empirical Processes in M-estimation},
  author={Geer, Sara A},
  volume={6},
  year={2000},
  publisher={Cambridge university press}
}

@article{qu2000improving,
  title={Improving generalised estimating equations using quadratic inference functions},
  author={Qu, Annie and Lindsay, Bruce G and Li, Bing},
  journal={Biometrika},
  volume={87},
  number={4},
  pages={823--836},
  year={2000},
  publisher={Oxford University Press}
}

@article{jiang2022usable,
  title={Usable and precise asymptotics for generalized linear mixed model analysis and design},
  author={Jiang, Jiming and Wand, Matt P and Bhaskaran, Aishwarya},
  journal={Journal of the Royal Statistical Society Series B: Statistical Methodology},
  volume={84},
  number={1},
  pages={55--82},
  year={2022},
  publisher={Oxford University Press}
}

@inproceedings{frechet1948elements,
  title={Les {\'e}l{\'e}ments al{\'e}atoires de nature quelconque dans un espace distanci{\'e}},
  author={Fr{\'e}chet, Maurice},
  booktitle={Annales de l'institut Henri Poincar{\'e}},
  volume={10},
  number={4},
  pages={215--310},
  year={1948}
}

@book{zuur2009mixed,
  title={Mixed effects models and extensions in ecology with R},
  author={Zuur, Alain F and Ieno, Elena N and Walker, Neil J and Saveliev, Anatoly A and Smith, Graham M and others},
  volume={574},
  year={2009},
  publisher={Springer}
}

@article{laird1982random,
  title={Random-effects models for longitudinal data},
  author={Laird, Nan M and Ware, James H},
  journal={Biometrics},
  pages={963--974},
  year={1982},
  publisher={JSTOR}
}

@article{matabuena2026beyond,
  title={Beyond scalar metrics: functional data analysis of postprandial continuous glucose monitoring in the AEGIS study},
  author={Matabuena, Marcos and Sartini, Joseph and Gude, Francisco},
  journal={BMC Medical Research Methodology},
  year={2026},
  publisher={Springer}
}

@article{bhattacharjee2023geodesic,
author = {Satarupa Bhattacharjee and Hans-Georg Müller},
title = {Geodesic Mixed Effects Models for Repeatedly Observed/Longitudinal Random Objects},
journal = {Journal of the American Statistical Association},
volume = {120},
number = {551},
pages = {1879--1892},
year = {2025},
publisher = {Taylor \& Francis},
doi = {10.1080/01621459.2025.2474267},


URL = { 
    
        https://doi.org/10.1080/01621459.2025.2474267
    
    

},
eprint = { 
    
        https://doi.org/10.1080/01621459.2025.2474267
    
    

}

}

@article{zhang2022nonlinear,
  author = {Qi Zhang and Lingzhou Xue and Bing Li},
  title = {Dimension Reduction for Fr{\'e}chet Regression},
  journal = {Journal of the American Statistical Association},
  year = {2024},
  volume = {119},
  number = {548},
  pages = {2733--2747},
  doi = {10.1080/01621459.2023.2277406}
}

@article{bhattacharjee2021single,
  author = {Satarupa Bhattacharjee and Hans-Georg M{\"u}ller},
  title = {Single Index Fr{\'e}chet Regression},
  journal = {The Annals of Statistics},
  year = {2023},
  volume = {51},
  number = {4},
  pages = {1770--1798},
  doi = {10.1214/23-AOS2307}
}

@article{qiu2024semisupervised,
  author = {Rui Qiu and Zhou Yu and Zhenhua Lin},
  title = {Semi-supervised Fr{\'e}chet Regression},
  journal = {arXiv preprint arXiv:2404.10444},
  year = {2024},
  doi = {10.48550/arXiv.2404.10444}
}

@article{schotz2021frechet,
  author = {Christof Sch{\"o}tz},
  title = {Nonparametric Regression in Nonstandard Spaces},
  journal = {Electronic Journal of Statistics},
  year = {2022},
  volume = {16},
  number = {2},
  pages = {4679--4741},
  doi = {10.1214/22-EJS2056}
}

@article{hanneke2022universally,
  author = {Steve Hanneke and Aryeh Kontorovich and Sivan Sabato and Roi Weiss},
  title = {Universal Bayes Consistency in Metric Spaces},
  journal = {The Annals of Statistics},
  year = {2021},
  volume = {49},
  number = {4},
  pages = {2129--2155},
  doi = {10.1214/20-AOS2029}
}

@article{bulte2023medoid,
  author = {Matthieu Bult{\'e} and Helle S{\o}rensen},
  title = {Medoid Splits for Efficient Random Forests in Metric Spaces},
  journal = {Computational Statistics \& Data Analysis},
  year = {2024},
  volume = {198},
  pages = {107995},
  doi = {10.1016/j.csda.2024.107995}
}

@article{bhattacharjee2023nonlinear,
  author = {Satarupa Bhattacharjee and Bing Li and Lingzhou Xue},
  title = {Nonlinear Global Fr{\'e}chet Regression for Random Objects via Weak Conditional Expectation},
  journal = {The Annals of Statistics},
  year = {2025},
  volume = {53},
  number = {1},
  pages = {117--143},
  doi = {10.1214/24-AOS2457}
}

\newpage

\end{document}